\documentclass{article}

\usepackage[english]{babel}

\usepackage[a4paper,top=2cm,bottom=2cm,left=3cm,right=3cm,marginparwidth=1.75cm]{geometry}

\usepackage{amsmath}
\usepackage{graphicx}
\usepackage[colorlinks=true, allcolors=blue]{hyperref}

\usepackage{hyperref}       
\usepackage{url}            
\usepackage{booktabs}       
\usepackage{amsfonts}       
\usepackage{nicefrac}       
\usepackage{microtype}      
\usepackage{lipsum}		
\usepackage{natbib}
\usepackage{doi}

\usepackage{subcaption}

\usepackage{amsmath,amsfonts}
\usepackage{algorithm}
\usepackage{algorithmicx}
\usepackage{algpseudocode}

\algnewcommand\algorithmicforeach{\textbf{for each}}
\algdef{S}[FOR]{ForEach}[1]{\algorithmicforeach\ #1\ \algorithmicdo}

\title{Enhancing Vehicle Environmental Awareness via Federated Learning and Automatic Labeling}
\author{Chih-Yu Lin and Jin-Wei Liang}

\begin{document}
\maketitle

\begin{abstract}
Vehicle environmental awareness is a crucial issue in improving road safety. Through a variety of sensors and vehicle-to-vehicle communication, vehicles can collect a wealth of data. However, to make these data useful, sensor data must be integrated effectively. This paper focuses on the integration of image data and vehicle-to-vehicle communication data. More specifically, our goal is to identify the locations of vehicles sending messages within images, a challenge termed the vehicle identification problem. In this paper, we employ a supervised learning model to tackle the vehicle identification problem. However, we face two practical issues: first, drivers are typically unwilling to share privacy-sensitive image data, and second, drivers usually do not engage in data labeling. To address these challenges, this paper introduces a comprehensive solution to the vehicle identification problem, which leverages federated learning and automatic labeling techniques in combination with the aforementioned supervised learning model. We have validated the feasibility of our proposed approach through experiments.
\end{abstract}

\section{Introduction}

Vehicle environmental awareness refers to a vehicle's ability to identify and understand its surroundings using various sensing technologies. It is a critical aspect of enhancing road safety and serves as the basis for autonomous driving technology. To enable vehicles to have a more precise understanding of their surroundings, a wide range of sensors are installed in vehicles, including radar, LiDAR, GPS, cameras, and more. Additionally, through vehicle-to-vehicle communication, vehicles can exchange information with each other, allowing them to better comprehend the intentions of other vehicles in their vicinity. However, to make these data useful, they must be fused.

\begin{figure}
    \centering
    \begin{subfigure}[b]{0.235\textwidth}
        \centering
        \includegraphics[width=\textwidth]{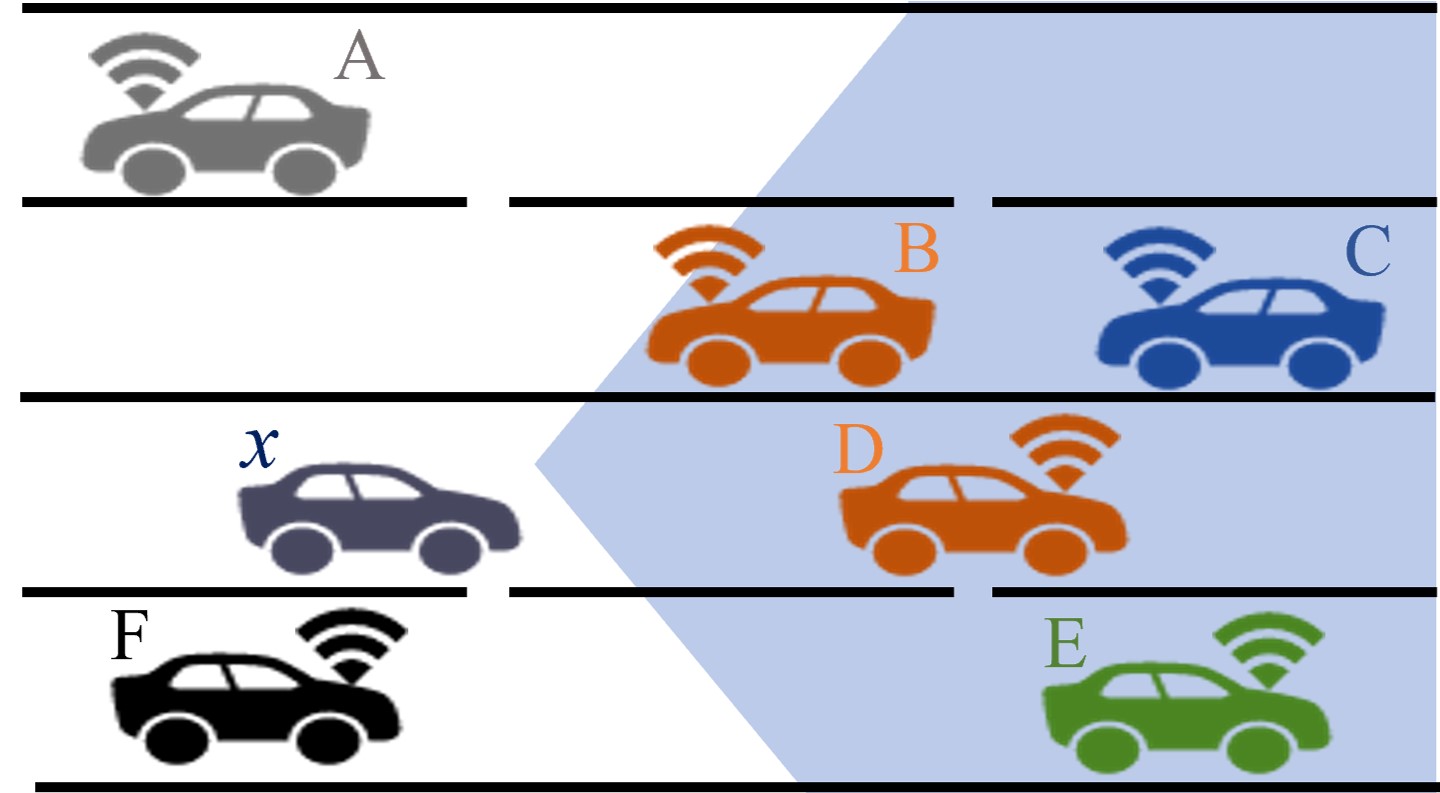}
    \end{subfigure}
    \begin{subfigure}[b]{0.235\textwidth}
        \centering
        \includegraphics[width=\textwidth]{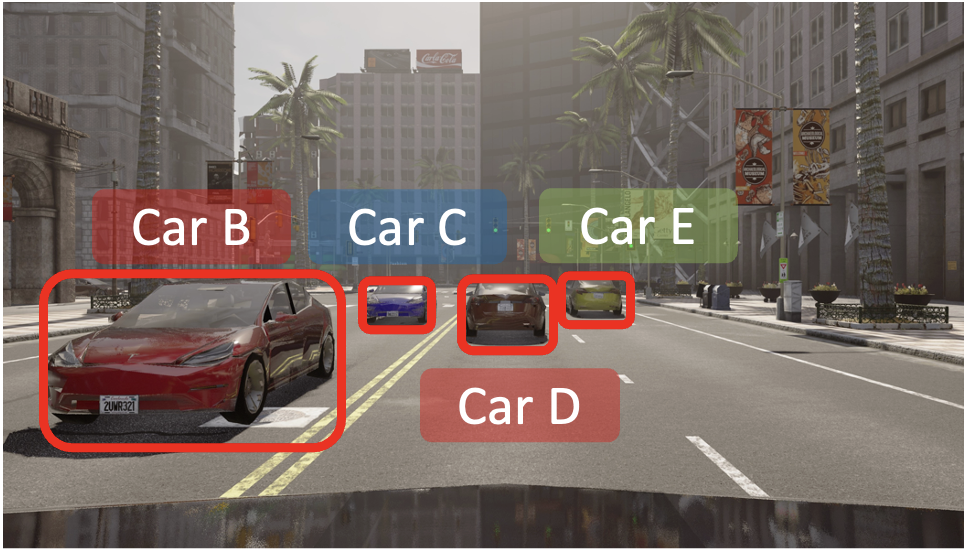}
    \end{subfigure}
    \caption{The vehicle identification (VID) problem.}
    \label{fig:vid_problem}
\end{figure}

Data fusion is the technology that merges data from various sources to create more comprehensive and valuable information. The purpose of data fusion is to gain deeper and more holistic insights than what can be achieved using data from a single source alone. In \cite{mdfnn2020}, an approach to vehicle environmental awareness is proposed, which involves the fusion of image data and vehicle-to-vehicle communication data. To illustrate the problem it addresses, consider Fig.~\ref{fig:vid_problem}. We assume that vehicle $x$ has a camera that captures images in front of vehicle. At the same time, vehicle $x$ can receive messages from other surrounding vehicles. The challenge is how vehicle $x$ can determine which vehicle in the image sends the message when vehicle $x$ receives a message. This problem is referred to as the Vehicle Identification (VID) problem. We have proposed a neural network model called Mapping Decision Feedback Neural Network (MDFNN) to address the VID problem\cite{mdfnn2020}.

However, MDFNN is a supervised model. This means that after collecting vehicle-to-vehicle communication data and image data, a person must manually pair the vehicle-to-vehicle communication data with the image data to generate sufficient training data to train the MDFNN model. It can be expected that drivers are typically unwilling to perform this labeling task. Therefore, a more viable approach is to upload all the data to the cloud and have dedicated individuals perform the labeling and model training. However, dealing with a large volume of data not only consumes human resources but also requires a significant amount of network bandwidth for data uploading. Additionally, as data carries a certain degree of privacy, uploading data to the cloud raises privacy concerns.

Federated learning \cite{fl_pmlr}\cite{fl_9084352}\cite{fl_10.1145/3298981} provides a potential solution to the privacy concerns mentioned above. Federated learning consists of numerous local nodes, also known as federated learning clients, each of which trains a local model based on its locally held data. Subsequently, all local nodes transmit their model parameters to a parameter server. The parameter server aggregates the parameters from these local models and constructs a global model. Then, this global model is returned to the local nodes. Since the global model is a combination of many local models, it is expected to perform better than a model trained by an individual local node. In this process, local nodes do not need to transmit sensitive data to the parameter server; they only need to send their local model parameters. As a result, data privacy is reasonably preserved.

We can apply federated learning to the aforementioned VID problem. For example, we can treat each vehicle as a local node, where vehicles only need to transmit model parameters to the parameter server, eliminating the need to send privacy-sensitive image data or vehicle-to-vehicle communication data to the server. However, when the model is a supervised model, the local nodes would typically require some labeled training data to train their local models. The challenge then becomes how to acquire labeled data. In the context of our in-vehicle network environment, expecting drivers to perform labeling is essentially impossible. Therefore, in the federated learning environment, local nodes sometimes need an automatic labeling method for their local data.

This paper presents a solution to the VID problem, which combines federated learning and an automatic labeling approach. We refer to this solution as FedMDFNN (Federated MDFNN). In FedMDFNN, we assume the presence of a federated learning parameter server in the cloud, with each vehicle regarded as a local node in the federated learning system. Each vehicle independently trains its local MDFNN based on its acquired data and uploads the relevant MDFNN parameters to the parameter server. The parameter server aggregates these parameters from all local nodes, resulting in new model parameters, which are then transmitted back to each vehicle. In FedMDFNN, each vehicle is not required to upload its private data, ensuring data privacy.

In addition, to address the challenge of labeling training data, this paper introduces an automatic labeling method using license plate recognition and data augmentation techniques. This automatic labeling approach makes FedMDFNN feasible. Finally, we have validated the proposed FedMDFNN through extensive experiments.

The remainder of this paper is organized as follows. Sect.~\ref{sec:relatedwork} discusses some related work. Sect.~\ref{sec:fedmdfnn} describes our proposed FedMDFNN. Sect.~\ref{sec:experiments} details our simulation setup and performance evaluations. 
Conclusions are drawn in Sect.~\ref{sec:conclusions}.

\section{Related Work}\label{sec:relatedwork}

\subsection{Vehicle environmental awareness}

Vehicle environmental awareness enables vehicles to understand and adapt to changes in road and environmental conditions. This relies on the real-time collection and processing of a vast amount of sensor data to identify key information such as road markings, obstacles, vehicles, pedestrians, etc.

Radar is a common automotive sensor that is used to detect the distance and speed of objects, such as other vehicles, obstacles, and pedestrians, in front of the vehicle \cite{radar_7485214,radar_9337403,radar_9216363}. Radar can operate effectively even in adverse weather conditions\cite{radar_9353210}. However, radar information alone cannot determine the intentions and driving states of vehicles\cite{radar_7535520}.

LiDAR (Light Detection and Ranging) is a sensor that uses laser beams to measure object positions and distances. It can generate precise three-dimensional models of the surrounding environment, including roads, buildings, and other obstacles\cite{lidar_8529992}. Like radar, LiDAR alone cannot determine the intentions and driving states of vehicles.

Cameras are another common automotive sensor that captures images of the road, including road signs, traffic signals, lanes, and pedestrians. Through image processing and recognition technologies, the system can analyze these images and identify important elements\cite{camera_9253415,camera_8910493,camera_9777323,camera_8668986}. Cameras can also be used to measure the distance and speed of vehicles in front, and even track vehicles\cite{camera_7759904,camera_8575384,camera_Xu2016EndtoEndLO}.

With the maturity of vehicle-to-vehicle communication technology, communication devices can also be considered as a type of sensor. Through the information transmitted by other vehicles, a vehicle can gain insight into the intentions and states of nearby vehicles, thus enhancing road safety\cite{camera_7835937}.


The various sensors mentioned above have their limitations when used individually. Therefore, we can achieve better vehicle environmental awareness by using multiple sensors\cite{fusion_8612054}. When using two or more sensors simultaneously, we have to address the issue of data fusion\cite{fusion_s23063335,fusion_9668695}. \cite{fusion_8598907} fuses radar sensors and communication data, while \cite{mdfnn2020} and \cite{fusion_8885487} fuse cameras and vehicle-to-vehicle communication. The VID problem addressed in this paper can be considered as the fusion of image data and communication data.

Lastly, with the rapid advancement of artificial intelligence technology, many studies have used AI to analyze sensor data and formulate optimal driving strategies\cite{ai_9266686,ai_9337402,ai_slpc,ai_9191134}.


\subsection{Federated Learning}

In traditional supervised machine learning, service providers need to collect user data and label the data to train a model. However, this approach can pose privacy risks. Additionally, transmitting large amounts of data to a server can consume significant bandwidth and time. Federated learning provides a solution to these problems\cite{fl_pmlr}. In fact, common issues in federated learning such as data heterogeneity and participant selection have also been studied\cite{fl_10153777, fl_MA2022244, fl_9714733}.

The research on applying federated learning to vehicular networks has also been explored\cite{fl_vehiclenetwork9205482}. An enhanced federated learning approach for Intelligent Transportation Systems (ITS) is presented in \cite{fl_9714733}. In \cite{fl_9685068}, a novel federated learning framework named FedVANET is introduced for vehicle ad hoc networks (VANET). FedVANET is capable of handling Non-IID data and can be applied to dynamic topologies. Federated learning frameworks have been proposed to tackle the challenge of trajectory prediction\cite{fl_vehiclenetwork10076689, fl_fedvanet_tp10322884}. However, the above-mentioned papers do not delve deeply into the data labeling issue for federated clients. 

The model proposed in this paper for the VID problem can be applied within a federated learning framework. Each vehicle can be regarded as a federated client, allowing each vehicle to avoid uploading sensitive image data. This not only ensures data privacy, but also saves the network bandwidth required for uploading image data. In particular, in this paper, we focus on discussing the data labeling issue for federated clients. In the assumed scenario, it is nearly impossible for vehicle drivers to perform labeling tasks. Therefore, we propose an automatic labeling approach to address the data labeling issue for federated clients. In the future, we will further consider the impact of data imbalance on the proposed method.

\subsection{Automatic Labeling}

Through federated learning, we address the issue of data privacy. However, when the model is a supervised model, we encounter the data labeling problem that is not discussed by most papers focusing on federated learning. In this scenario, we may need to complement with an automatic labeling method.

\cite{al_Zhang2021ASO} provides a review of automatic labeling methods for video, audio, and text data. \cite{al_9778081} proposes a clustering method based on Gauss chaotic mapping particle swarm optimization (GCMPSO) to improve the accuracy of automatically labeling human activities. \cite{al_9564211} presents an automatic labeling approach for automatically labeling daily activities, such as walking and driving behaviors.\cite{al_9289591} introduces an automatic labeling approach for labeling Synthetic Aperture Radar (SAR) images, using GPS data and open APIs supplied by government agencies. \cite{al_9150584} trains a model to filter automatically labeled data and uses the filtered data to train the final model, resulting in improved model performance.

In this paper, considering the assumed scenario where it is impractical for drivers to manually label data, we propose an automatic labeling method that combines license plate recognition with data augmentation. To the best of our knowledge, this paper is the first to propose an automatic labeling approach for the VID problem. With the introduction of automatic labeling, our proposed method eliminates the need for drivers to perform labeling tasks, making the overall approach more feasible.


\section{FedMDFNN}\label{sec:fedmdfnn}

The VID problem addressed in this paper can be defined as follows. We assume that vehicle $x$ is equipped with a front-view camera. Through this camera, vehicle $x$ can use object recognition technology to find the bounding boxes of vehicles captured by the front-view camera. These vehicles are represented as $V^{t}=\{v^{t}_{i} \mid 1 \le i \le n\}$, where $t$ represents the time when the image is captured, $n$ represents the number of vehicles recognized by object recognition technology, and $v^{t}_{i}$ represents vehicle $i$ appearing in the capture range of the front-view camera of vehicle $x$ at time $t$. In addition, we assume that vehicle $x$ has received $m$ messages at time $t$, represented as $B^{t}=\{b^{t}_{j} \mid 1 \le j \le m\}$, where $b^{t}_{j}$ represents the message sent by vehicle $j$. The task of the VID problem is to find the pairing of $V^t$ and $B^t$, represented as $P^{t}=\{(v^{t}_{i}, b^{t}_{j}) \mid v^{t}_{i} \in V^{t} \land b^{t}_{j} \in B^t\}$.

\begin{figure}
    \centering
    \includegraphics[width=8.8cm]{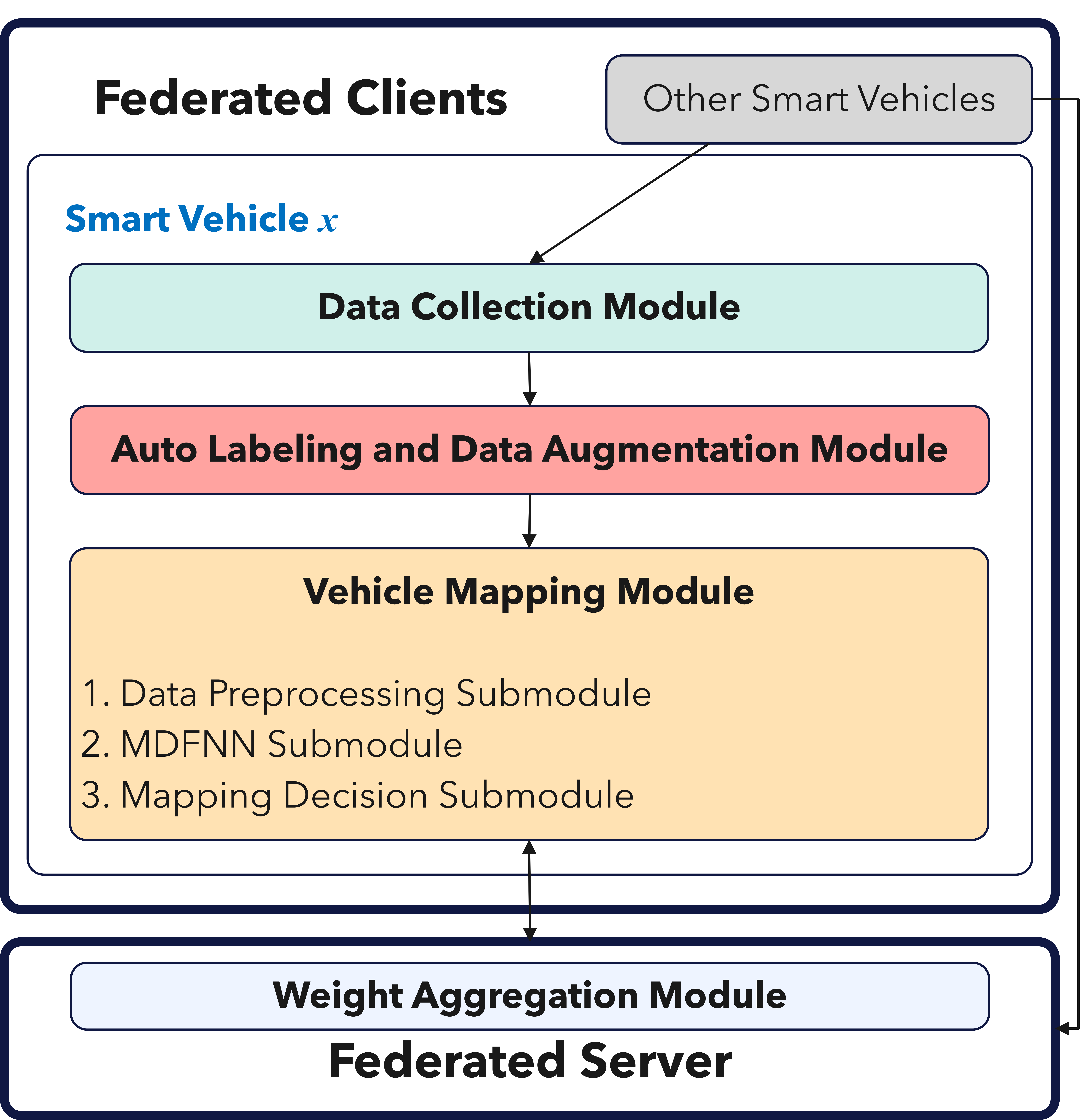}
    \caption{System architecture of FedMDFNN.}
    \label{fig:fedmdfnn}
\end{figure}

To address the VID problem, we propose a solution called FedMDFNN (Federated MDFNN), which is based on federated learning and automatic labeling. The system architecture of FedMDFNN is illustrated in Fig.~\ref{fig:fedmdfnn}. Following the architecture of federated learning, FedMDFNN is divided into two components: the federated server and the federated clients. The task of a federated server is to aggregate the model parameters from federated clients to build a global model. Federated clients represent participating vehicles and consist of three modules: (1) data collection module, (2) automatic labeling and data augmentation module, and (3) vehicle mapping module.

Firstly, the data collection module collects image data from the front-view camera and the rear-view camera of vehicle $x$. Through front-view camera and object recognition technology, vehicle $x$ can acquire $V^{t}$. The rear-view camera is used for the data augmentation submodule, as described below. Furthermore, through vehicle-to-vehicle communication, vehicle $x$ obtains $B^{t}$. To perform the pairing, vehicle $x$ must also collect data from its own sensors, denoted by $S^{t}$. The details of the data collection module are described in Sect.~\ref{subsec:datacollection}.

Then, the automatic labeling and data augmentation module is utilized to generate labeled data. This module comprises two submodules: the automatic labeling submodule and the data augmentation submodule. The automatic labeling submodule attempts to establish initial pairings for $V^{t}$ and $B^{t}$ through license plate recognition. This initial pairing is denoted as $P^{t}_{auto}$, where $P^{t}_{auto} \subseteq P^{t}$. On the other hand, the data augmentation submodule aims to identify vehicles outside the horizontal field of view of the front-view camera but whose messages have been received by vehicle $x$, represented by $V^{t}_{Outside}=\{v^{t}_{i} \mid v^{t}_{i} \notin V^{t} \land b(v^{t}_{i}) \in B^t\}$, where $b(v^{t}_{i})$ signifies the message sent by vehicle $v^{t}_{i}$. The specifics are elaborated in Sect.~\ref{subsec:automaticlabeling}.


Finally, the vehicle mapping module primarily consists of a neural network model. It combines $P^{t}_{auto}$ and $V^{t}_{Outside}$ to generate training data for the model. Once the model is trained, $S^{t}$ and $B^{t}$ serve as inputs, and the output of the model is the predicted positions of the vehicles in the image. Lastly, through the mapping decision submodule, $B^{t}$ and $V^{t}$ are paired. The details are explained in Sect.~\ref{subsec:mappingmodule}.

Table~\ref{tab:notation} summarizes the notation used in the paper.

\begin{table}
    \centering
    \caption{Notation}
    \begin{tabular}{| c | l |}
        \hline
        \textbf{Symbol} & \textbf{Description} \\        
        \hline
        $V^{t}$ & \multicolumn{1}{p{6.6cm}|}{The set of vehicles captured by the front-view camera of vehicle $x$ at time $t$ and recognized by object recognition software, including their bounding boxes.}\\
        \hline
        $B^{t}$ & \multicolumn{1}{p{6.6cm}|}{The set of messages from other vehicles received by vehicle $x$ at time $t$.}\\
        \hline
        $P^{t}$ & The vehicle pairing set.\\
        \hline
        $S^{t}$ & The relevant sensor values of vehicle $x$ at time $t$.\\
        \hline
        $P^{t}_{auto}$ & \multicolumn{1}{p{6.6cm}|}{The set of automatic pairings using license plate recognition technology at time $t$.}\\
        \hline
        $V^{t}_{Outside}$ & \multicolumn{1}{p{6.6cm}|}{The set of vehicles at time $t$ that did not appear in the front-view camera of vehicle $x$ but whose messages were received by vehicle $x$.}\\
        \hline
        $V^{t}_{Rear}$ & \multicolumn{1}{p{6.6cm}|}{The set of vehicles captured by the rear-view camera of vehicle $x$ at time $t$ and identified using license plate recognition.}\\
        \hline
    \end{tabular}
    \label{tab:notation}
\end{table}

\subsection{Data Collection Module}\label{subsec:datacollection}

The data collection module collects image data from the front-view camera and the rear-view camera of vehicle $x$. Through the front-view camera and object recognition technology, we can obtain the collection of vehicle bounding boxes $V^{t}$. The bounding box coordinates for each vehicle $v^{t}_{i}$ in $V^{t}$ are normalized and can be represented as $v^{t}_{i}.bb\_norm = [\frac{v^{t}_{i}.x_{TopLeft}}{w}, \frac{v^{t}_{i}.y_{TopLeft}}{h}, \frac{v^{t}_{i}.x_{BottomRight}}{w}, \frac{v^{t}_{i}.y_{BottomRight}}{h}]$, where $w$ and $h$ represent the width and height of the image, respectively. The image data from the rear-view camera is utilized for the data augmentation submodule, and the details are explained in Sect.~\ref{subsubsec:dataaugmentation}.

On the other hand, through vehicle-to-vehicle communication, vehicle $x$ can receive communication data $b_y^t$, from other vehicles, with vehicle $y$ as an illustrative example. These data include the following:
\begin{itemize}
\item GPS data of vehicle $y$ at time $t$, represented by $v_y^t.lat$ and $v_y^t.lng$.
\item Orientation sensor data of vehicle $y$ at time $t$, represented by $v_y^t.ori$.
\item Speed of vehicle $y$ at time $t$, represented by $v_y^t.spd$.
\item ID of vehicle $y$, represented by $v_y^t.id$. For privacy concerns, $v_y^t.id$ is obtained by hashing the license plate of vehicle $y$.
\item State of vehicle $y$, which enables vehicle $x$ to comprehend the driving conditions of surrounding vehicles.
\end{itemize}
All messages received from time $t-1$ to time $t$ constitute the set $B^t$.

To facilitate pairing, vehicle $x$ must also collect data from its own sensors, represented as $S^t=(v_x^t.lat, v_x^t.lng, v_x^t.ori,\\ v_x^t.spd)$. The VID problem can be viewed as matching $V^t$ and $B^t$ using $S^t$.

\subsection{Auto Labeling and Data Augmentation Module}\label{subsec:automaticlabeling}

The automatic labeling and data augmentation module aims to automatically generate labeled training data. The automatic labeling submodule uses license plate recognition technology to establish pairings for $V^t$ and $B^t$, creating a pairing set $P_{auto}^t$, where $P_{auto}^t \subseteq P^t$. Simultaneously, the data augmentation submodule identifies vehicles outside the horizontal field of view of the front-view camera of vehicle $x$ (i.e., not in $V^t$), but whose messages have been received by vehicle $x$, denoted as $V_{Outside}^t=\{v_i^t \mid v_i^t \notin V^t \land b(v_i^t) \in B^t\}$. The purpose of both submodules is to generate training data, and the details are described separately.

\subsubsection{Auto Labeling Submodule}

The auto labeling submodule uses license plate recognition technology to match $V^t$ and $B^t$, aiming to establish a pairing set $P_{auto}^t$. The approach involves attempting to identify the license plate bounding box for $v_i^t\in V^t$ and using Optical Character Recognition (OCR) technology to extract the license plate number. Then, the extracted license plate number is hashed by a common hash function and compared with the vehicle ID carried by the transmitted packet, facilitating the matching of $V^t$ and $B^t$.

\begin{figure}
    \centering
    \begin{subfigure}[b]{0.23\textwidth}
        \centering
        \includegraphics[width=\textwidth]{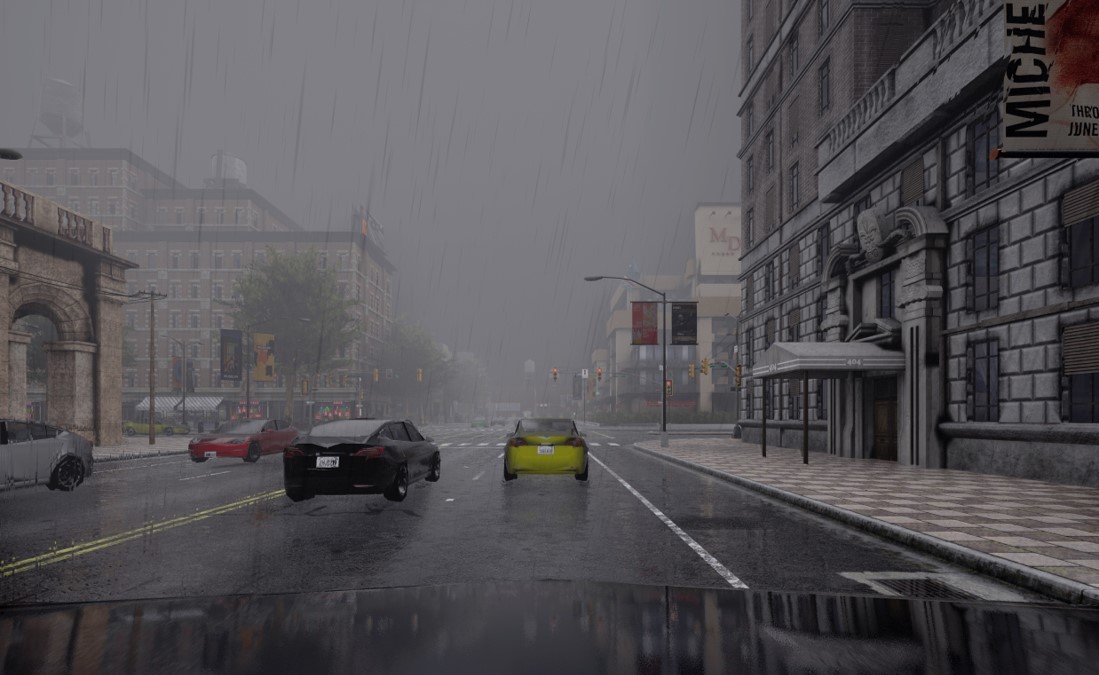}
        \caption{OCR failed due to bad weather.}
    \end{subfigure}
    \begin{subfigure}[b]{0.23\textwidth}
        \centering
        \includegraphics[width=\textwidth]{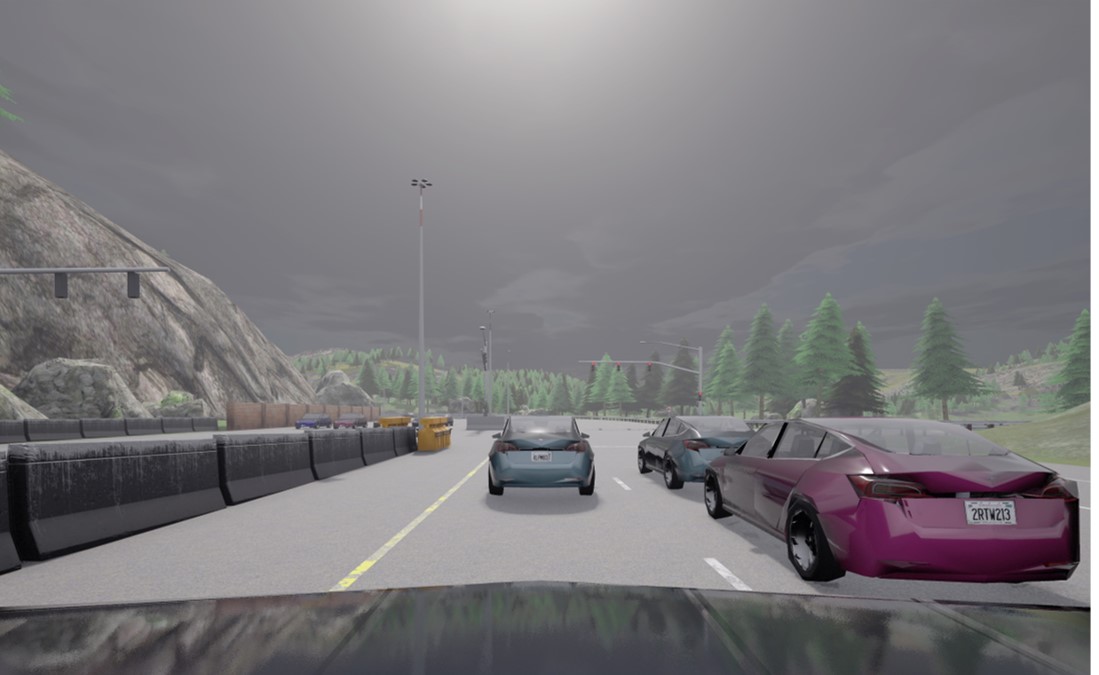}
        \caption{OCR failed due to license plate being obscured.}
    \end{subfigure}
    \caption{Examples of OCR failure.}
    \label{fig:ocrfail}
\end{figure}

It should be noted that due to various factors, such as different weather conditions, distances, and the state of the preceding vehicles, not all vehicles in $V^{t}$ can be successfully matched using license plate recognition. For example, Fig.~\ref{fig:ocrfail}(a) illustrates a case where license plates cannot be successfully identified due to bad weather, and Fig.~\ref{fig:ocrfail}(b) shows a case where license plates cannot be recognized due to being obstructed by other vehicles. Therefore, the use of license plate recognition technology can only achieve partial matching, and the remaining vehicles can be matched using the FedMDFNN proposed in this paper.

We utilize the OCR function provided by the Google Cloud Vision API\cite{googlecloudvisionapi} to implement license plate recognition. Additionally, we employ a confusing character table, established through experiments, to refine the OCR recognition results. The reason for creating this table is that, during the license plate recognition process, we observed that certain confusing characters can affect the accuracy of recognition. For example, OCR software often confuses the number 0 with the English letter O or misinterprets the number 1 as the English letter I. Relying solely on an exact matching approach for character comparison would lead to the inability to automatically label a significant amount of data.

\begin{figure}
    \centering
    \includegraphics[width=8cm]{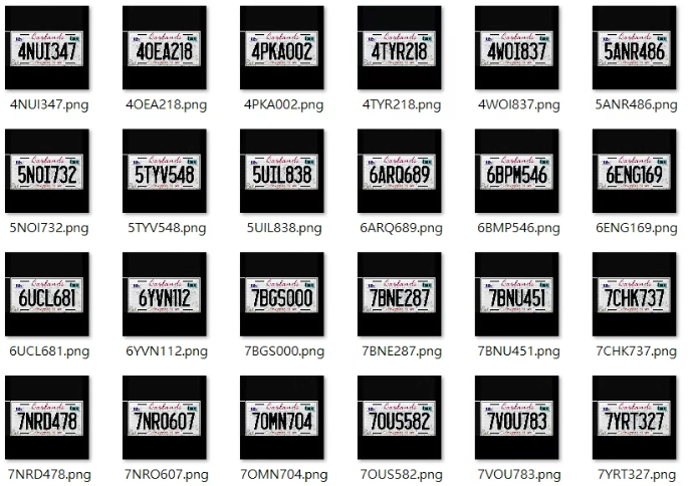}
    \caption{Built-in license plates in the Carla simulator\cite{carla_Dosovitskiy17}.}
    \label{fig:carlalps}
\end{figure}

Below is an example that illustrates the process of establishing a confusing character table through experiments. We extracted 117 built-in license plates from the Carla simulator\cite{carla_Dosovitskiy17}, as depicted in Fig.~\ref{fig:carlalps}. To simulate real-world conditions more accurately, we adjusted the pixel dimensions of the license plates to $86 \times 41$ pixels, closely resembling the dimensions of license plates captured by dashcams. Based on the recognition results for these 117 license plates, we obtained the confusing character table, presented in Table~\ref{tab:cctable}. This table displays the number of occurrences and the percentage of OCR results.

\begin{table}
    \centering
    \caption{The confusing character table example.}
    \begin{tabular}{| c | l |}
        \hline
        \textbf{Character} & \textbf{OCR Result: the number of occurrences (percentage)} \\        
        \hline
        0 & 0: 20 (61\%), \textbf{O: 13 (39\%)}\\
        \hline
        1 & 1: 42 (84\%), T: 1 (2\%), E: 1 (2\%), I: 6 (12\%)\\
        \hline
        2 & 2: 40 (91\%), Z: 4 (9\%)\\
        \hline
        3 & 3: 39 (95\%), L: 1 (2\%), 6: 1 (2\%)\\
        \hline
        4 & 4: 50 (100\%)\\
        \hline
        5 & 5: 39 (95\%), S: 2 (5\%)\\
        \hline
        6 & 6: 30 (88\%), W: 1 (3\%), 3: 1 (3\%), G: 2 (6\%)\\
        \hline
        7 & 7: 62(95\%), A: 1(2\%), T: 2(3\%) \\
        \hline
        8 & 8: 66(96\%), B: 3(4\%) \\
        \hline
        9 & 9: 40(100\%) \\
        \hline
        A & A: 19(100\%) \\
        \hline
        B & B: 12(86\%), L: 1(7\%), 9: 1(7\%) \\
        \hline
        C & C: 14(100\%) \\
        \hline
        D & D: 5(71\%), \textbf{0: 2(29\%)} \\
        \hline
        E & E: 21(95\%), C: 1(5\%) \\
        \hline
        F & F: 8(100\%) \\
        \hline
        G & G: 7(100\%) \\
        \hline
        H & H: 16(100\%) \\
        \hline
        I & I: 12(75\%), \textbf{1: 4(25\%)} \\
        \hline
        J & J: 12(100\%) \\
        \hline
        K & K: 17(100\%) \\
        \hline
        L & L: 20(100\%) \\
        \hline
        M & M: 18(85\%), 9: 1(5\%),  P: 1(5\%), H: 1(5\%) \\
        \hline
        N & N: 31(100\%) \\
        \hline
        O & O: 8(42\%), \textbf{0: 11(58\%)} \\
        \hline
        P & P: 8(89\%), M: 1(11\%) \\
        \hline
        Q & Q: 1(33\%), \textbf{0: 2(67\%)} \\
        \hline
        R & R: 22(100\%) \\
        \hline
        S & S: 2(22\%), \textbf{5: 7(78\%)} \\
        \hline
        T & T: 13(93\%), 7: 1(7\%) \\
        \hline
        U & U: 17(100\%) \\
        \hline
        V & V: 10(91\%), W: 1(9\%) \\
        \hline
        W & W: 4(80\%), M: 1(20\%) \\
        \hline
        X & X: 7(100\%) \\
        \hline
        Y & Y: 9(100\%) \\
        \hline
        Z & Z: 3(100\%) \\
        \hline
    \end{tabular}
    \label{tab:cctable}
\end{table}

Through Table~\ref{tab:cctable}, we can identify characters with error rates that exceed the threshold $T$. Taking the letter 'I' as an example, it appears 16 times on the 117 license plates. Among these occurrences, it is misrecognized as the numeric character '1' four times. Assuming $T$ is set to $0.2$, as $\frac{4}{16} > 0.2$, we consider the letter 'I' and the digit '1' as a pair. This process generates a character pairing set $CP = \{(c_1, c_2) \mid err(c_1,c_2) > T \lor err(c_2,c_1) > T\}$, where $err(c_1,c_2)$ represents the probability of character $c_1$ being recognized as $c_2$. With the character pairing set $CP$, we can create a key-value table named character conversion table. The algorithm for building the character conversion table is shown in Algo.~\ref{algo:characterconvrsiontable}.

\begin{algorithm}
    \caption{Character Conversion Table Generator}\label{algo:characterconvrsiontable}
    \begin{algorithmic}[1]
    \State $CCT = \emptyset$
    \State $value = 1$
    \ForEach{$(c_1, c_2) \in CP$}
        \If{$CCT.hasKey(c_1)$}
            \State $CCT = CCT \cup (c_2, CCT.getValue(c_1))$
        \ElsIf{$CCT.hasKey(c_2)$}
            \State $CCT = CCT \cup (c_1, CCT.getValue(c_2))$
        \Else
            \State $CCT = CCT \cup (c_1, value)$
            \State $CCT = CCT \cup (c_2, value)$
            \State $value = value + 1$            
        \EndIf
    \EndFor
    \end{algorithmic}
\end{algorithm}

For a pair $(c_1, c_2)$ in $CP$, if either character is already in the character conversion table, the other character is added to the table using the same value. If neither character is present in the character conversion table, a new value is assigned. Based on the confusing character table (Table~\ref{tab:cctable}), the character pairing set $CP$, and the character conversion table generator algorithm, we obtain Table~\ref{tab:charconversiontable}.

\begin{table}
    \centering
    \caption{Character Conversion Table}
    \begin{tabular}{| c | l |}
        \hline
        \textbf{Key} & \textbf{Value} \\        
        \hline
        0 & \#1\\
        \hline
        O & \#1\\
        \hline
        D & \#1\\
        \hline
        Q & \#1\\
        \hline
        1 & \#2\\
        \hline
        I & \#2\\
        \hline
        5 & \#3\\
        \hline
        S & \#3\\
        \hline
    \end{tabular}
    \label{tab:charconversiontable}
\end{table}

Through the character conversion table, when a character from a license plate appears in the list of keys, we replace that character with its corresponding value. For example, suppose that we receive a message with the license plate number `5CRD321.' According to Table~\ref{tab:charconversiontable}, this license plate number is transformed into `\#3CR\#132\#2.' Consequently, even if the OCR recognition software mistakenly interprets it as `SCRO32I,' with the transformed license plate `\#3CR\#132\#2,' we can still successfully match it. To distinguish between converted and unconverted numeric characters, we prepend a `\#' symbol to the converted values. In this way, for example, even after conversion, we can correctly distinguish between `1ABCEF' and `2ABCEF'. Finally, as mentioned earlier, the vehicle's ID is generated by the license plate number and a hash function. Before performing the hash calculation, it is necessary to convert the license plate number based on Table~\ref{tab:charconversiontable} so that the receiving vehicles can make comparisons.

According to the license plate recognition method described above, we can match $V^{t}$ and $B^{t}$, and establish an initial pairing set $P_{auto}^t$. As mentioned above, not all vehicles in $V^{t}$ can successfully undergo pairing using the license plate recognition method. Therefore, an issue for discussion is what proportion of vehicles in $V^{t}$ can be successfully matched using this method. If this proportion is high, it implies that a simple license plate recognition method is sufficient for pairing, and there may not be a need for a complex vehicle mapping model. On the contrary, if this proportion is low, the training dataset that we can generate might be insufficient. Based on our simulation experiments, we find that the license plate recognition method can successfully match approximately 24\% of vehicles in a single photo.

\subsubsection{Data Augmentation Submodule}\label{subsubsec:dataaugmentation}

The purpose of the data augmentation module is to identify vehicles, denoted as $V^{t}_{Outside} = \{v^{t}_{i} \mid v^{t}_{i} \notin V^{t} \land b(v^{t}_{i}) \in B^t\}$, that are not in $V^{t}$ but have received their messages.

\begin{figure}
    \centering
    \includegraphics[width=8cm]{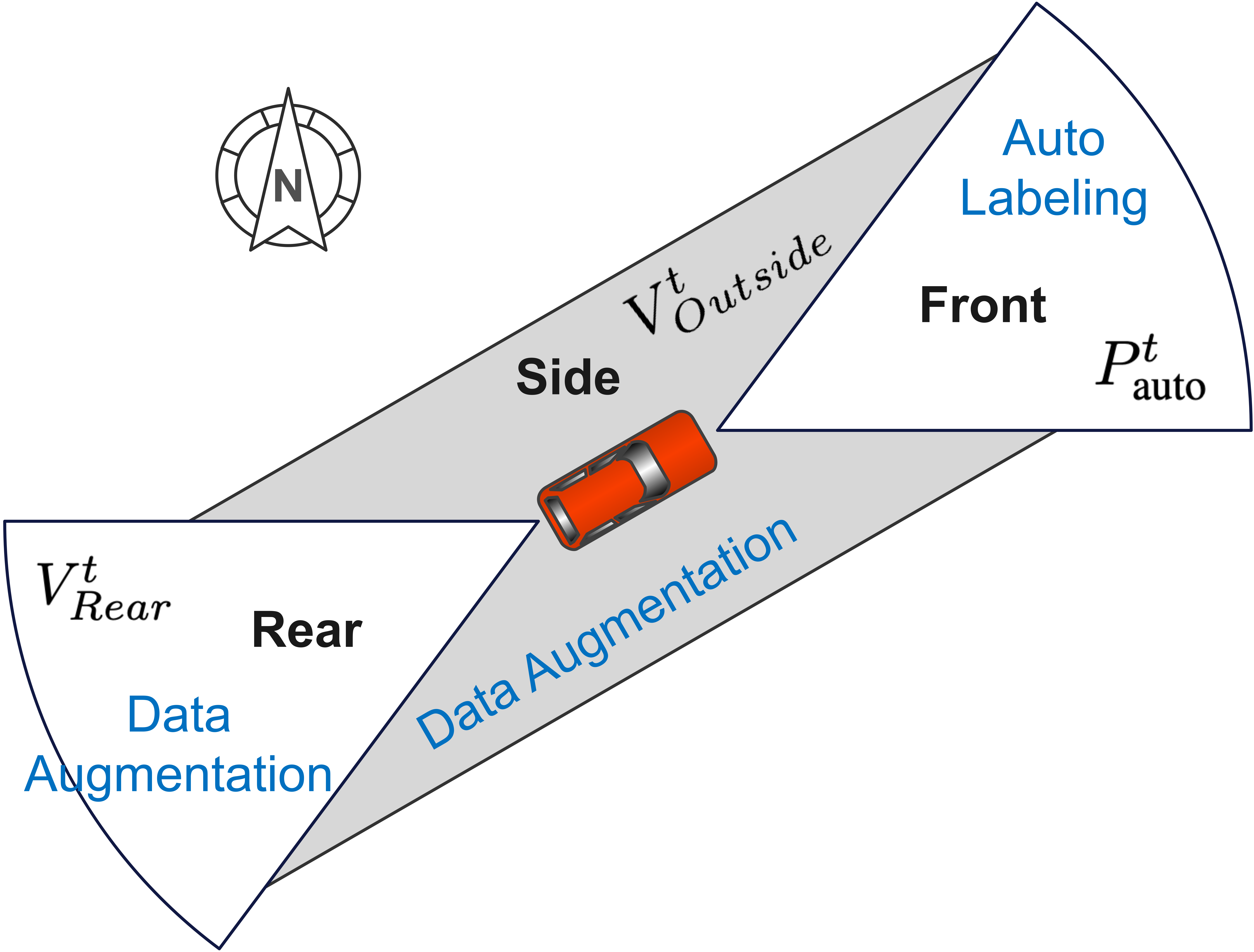}
    \caption{The front, rear, and sides of the vehicle. $P_{auto}^t$ is acquired through the automatic labeling submodule, while $V_{Rear}^t$ and $V_{Outside}^t$ are obtained via the data augmentation submodule.}
    \label{fig:dataaugmentation}
\end{figure}

We employ two methods for data augmentation. The first method involves using a rear-view camera. Assuming that the fields of view of the front-view and rear-view cameras do not overlap, we can apply the license plate recognition method mentioned earlier to recognize images captured by the rear-view camera. This results in $P^{t}_{Rear}=\{(v^{t}_{i}, b^{t}_{j}) \mid v^{t}_{i} \in V^{t}_{Rear} \land b^{t}_{j} \in B^{t}\}$, where $V^{t}_{Rear}$ represents vehicles that are captured by the rear-view camera (i.e., the vechicles located in the rear area in Fig.~\ref{fig:dataaugmentation}) and can be recognized using the license plate recognition method.

The second method of data augmentation considers vehicles on the sides, as illustrated in the gray areas in Fig.~\ref{fig:dataaugmentation}. For vehicle $x$, we can determine whether vehicle $v$ is within the horizontal field of view of $x$'s front-view camera using the following formula:
\begin{equation}\label{eq:horizontalfieldofview}
x.ori-\frac{\alpha}{2} \leq x.loc.bearing(v.loc) \leq x.ori+\frac{\alpha}{2},
\end{equation}
where $\alpha$ denotes the horizontal angle of view of the camera, $x.ori$ denotes the orientation of vehicle $x$, and $loc$ denotes the location obtained from GPS. If the formula holds, it indicates that vehicle $v$ is within the horizontal field of view of vehicle $x$'s camera. The set of vehicles that we aim to collect consists of those outside the horizontal field of view of the $x$'s front-view camera, but their messages are received by $x$, denoted as $V^{t}_{Outside}=\{v^{t}_{i} \mid v^{t}_{i} \notin V^{t} \land b(v^{t}_{i}) \in B^t\}$. It should be noted that, considering the inherent errors in GPS, we refer to the past $k$ seconds of data for vehicle $v$. If during these $k$ seconds, vehicle $v$ is consistently outside the horizontal field of view of vehicle $x$'s front-view camera, only then do we conclude that vehicle $v$ is indeed not within the horizontal field of view of the front-view camera of vehicle $x$.

In the end, we include the vehicles in $V^{t}_{Rear}$ for which vehicle $x$ has received their messages into $V^{t}_{Outside}$ to obtain the final $V^{t}_{Outside}$. It is important to note that the vehicles in $V^{t}_{Rear}$ cannot be part of the training data if vehicle $x$ has not received their messages. Additionally, there is a high probability that $V^{t}_{Rear}$ is a subset of $V^{t}_{Outside}$ (i.e., $V^{t}_{Rear} \subseteq V^{t}_{Outside}$). Therefore, if vehicles lack a rear-view camera and rely solely on GPS coordinates for orientation, we can still obtain $V^{t}_{Outside}$. The vehicles in $V^{t}_{Outside}$ constitute part of the training data.

\subsection{Vehicle Mapping Module}\label{subsec:mappingmodule}

As mentioned above, some vehicles cannot be successfully matched through license plate recognition, and the vehicle mapping module is designed for these cases. In our previous work, we have proposed a model called MDFNN for vehicle mapping\cite{mdfnn2020}. In this paper, we have made minor modifications to MDFNN. More importantly, compared to our previous work, there are two major differences:

\begin{itemize}
    \item In previous work, training data labeling was done manually. In this paper, training data are generated through the auto labeling and data augmentation module, enhancing the feasibility of the proposed solution.
    \item In this paper, we have incorporated federated learning, where vehicles do not need to upload images and communication data with privacy concerns. This ensures the privacy of the data and reduces the substantial bandwidth required for uploading image data.
\end{itemize}

For a neural network model, input data typically undergo preprocessing, including normalization. Sect.~\ref{subsubsec:datapreprocessing} details the data preprocessing method. The design of MDFNN is described in Sect.~\ref{subsubsec:mdfnn}, and the mapping decision submodule, responsible for final pairing, is explained in Sect.~\ref{subsubsec:mappingdecision}.

\subsubsection{Data Preprocessing Submodule}\label{subsubsec:datapreprocessing}

Sensors that can be used to determine the position of vehicles in the image include GPS sensors, speedometers, and orientation sensors. However, preprocessing of the sensor data is necessary to facilitate model training. The data preprocessing submodule handles the latitude and longitude information from GPS, along with the values from the orientation sensor and the speedometer, as follows.

First, the latitude and longitude values are converted to the decimal degrees format. In the longitude part, east longitude is represented as positive, while west longitude is represented as negative. In the latitude part, north latitude is represented as positive, and south latitude is represented as negative. After conversion, the difference between the latitude and longitude positions obtained from the received message and the vehicle's own latitude and longitude positions is calculated. This difference is then transformed into a value between $-1$ and $1$. For example, if the GPS signal extracted from the message after conversion is (+23.973875, +120.982025), and the vehicle's GPS signal after conversion is (+23.973828, +120.982038), then the difference is (0.000047, -0.000013). Finally, we normalize the latitude and longitude differences separately to ensure that they fall within the range of $-1$ and $1$.

\begin{figure}
    \centering
    \includegraphics[width=5.8cm]{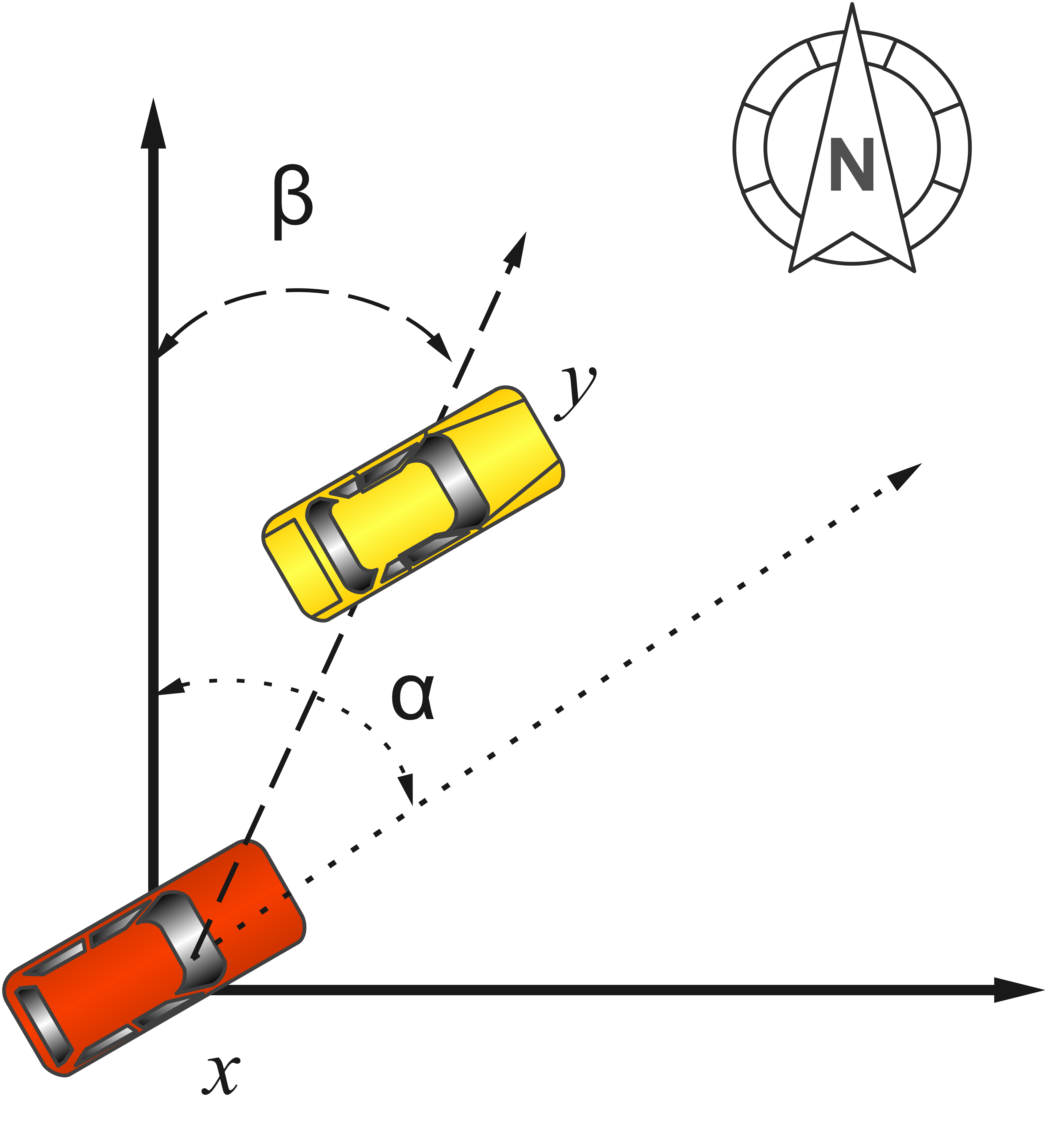}
    \caption{Illustration of orientation conversion method.}
    \label{fig:carorientation}
\end{figure}

The orientation sensor value also undergoes a conversion process, as depicted in Fig.~\ref{fig:carorientation}. Assume that the angle obtained by vehicle $x$ through its own orientation sensor is $\alpha$, and $\beta$ is calculated using the GPS information of $x$ and $y$. Through the following formula, we can calculate $\gamma$:
\begin{equation}\label{eq:processedorientationgamma}
\gamma =
\begin{cases}
    (\alpha-\beta)/180, & 180 \geq \alpha-\beta \geq -180\\
    (\alpha-\beta+360)/180, &  \alpha-\beta < -180\\
    (\alpha-\beta-360)/180, & \alpha-\beta > 180
\end{cases}
\end{equation}
The value of $\gamma$, if positive, indicates that vehicle $y$ is on the left side of vehicle $x$; conversely, if the value of $\gamma$ is negative, it indicates that vehicle $y$ is on the right side of vehicle $x$. Furthermore, the value of $\gamma$ ranges from $-1$ to $1$.

Finally, the vehicle speed collected by the speedometer is normalized to fall between $0$ and $1$.

\subsubsection{MDFNN Submodule}\label{subsubsec:mdfnn}

In our previous work, we proposed two types of MDFNN: the Grid-based MDFNN (Grid-MDFNN) and the Bounding box-based MDFNN (BBX-MDFNN)\cite{mdfnn2020}. We have made slight modifications to the design of the MDFNN, and here, we only provide a brief overview of the BBX-MDFNN.

\begin{figure}
    \centering
    \includegraphics[width=8.8cm]{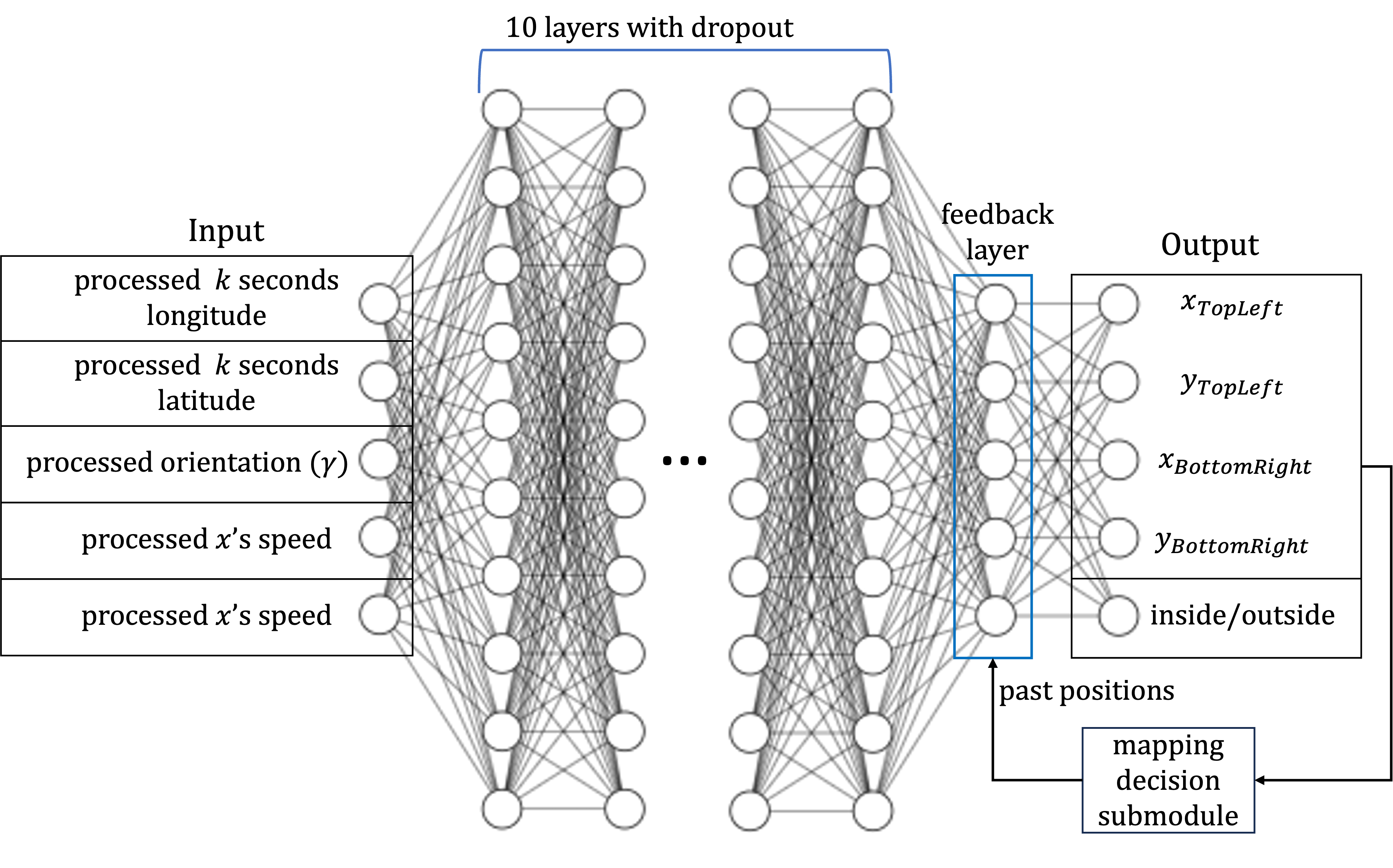}
    \caption{The neural network architecture of BBX-MDFNN.}
    \label{fig:bbx_mdfnn_nnarchitecture}
\end{figure}

The neural network architecture of BBX-MDFNN is depicted in Fig.\ref{fig:bbx_mdfnn_nnarchitecture}. The inputs include processed latitude and longitude values from the past $k$ seconds, speeds of vehicle $y$ (message sender) and vehicle $x$ (message receiver), and the processed orientation value $\gamma$. The output is the estimated bounding box of vehicle $y$ in the image, represented by a vector of length $5$ denoted as $O_{bbx}$. The first four elements ($O^1_{bbx}$ to $O^4_{bbx}$) of the vector represent the predicted bounding box position, while the last element ($O^5_{bbx}$) indicates whether vehicle $y$ is present in the image. The model consists of 10 hidden layers, and to prevent overfitting during training, we apply a 30\% dropout. Preceding the output layer, there is a feedback layer where the bounding boxes of vehicle $y$, calculated by the mapping decision module (explained in Section\ref{subsubsec:mappingdecision}), are input to collaboratively determine the final output.

The loss function used by BBX-MDFNN is as follows:
\begin{equation}\label{eq:bbx_mdfnn_loss}
Loss_{bbx} = \frac{1}{4} \sum_{i=1}^{4} (y_i - \hat{y}_i)^2 + \mu (y_{inside} - \hat{y}_{inside})^2,
\end{equation}
where $y_i$ represents the ground truth of the bounding box, $\hat{y}_i$ represents the predicted bounding box by BBX-MDFNN, $y_{inside}$ represents the ground truth of whether the vehicle $y$ is inside the image, $\hat{y}_{inside}$ represents the output value of BBX-MDFNN for whether the vehicle is inside the image, and $\mu$ is an adjustable weight.

Finally, we explain how the automatically generated training data are labeled in the BBX-MDFNN. For vehicles $v$ in $P_{auto}^{t}$, we use the bounding box ($v.bb_norm$) marked by the image recognition software for vehicle $v$ as labeled data, with $O^5_{bbx}$ set to $1$. For vehicles in $V_{outside}^{t}$, all five elements in $O_{bbx}$ are set to $0$.

\subsubsection{Mapping Decision Submodule}\label{subsubsec:mappingdecision}

We assume that for the vehicles in $B^t$, the set of bounding boxes estimated by BBX-MDFNN is denoted as $E^t$. Therefore, the original VID problem that pairs between $B^t$ and $V^t$ is transformed into the problem that pairs between $E^t$ and $V^t$. $E^t$ and $V^t$ may not match perfectly, and the number of vehicles in $E^t$ and $V^t$ may also be inconsistent. Therefore, we need a mapping decision submodule to determine the final pairing.

Algo.~\ref{algo:mappingdecision} outlines the mapping decision algorithm, which can be explained in four parts.

\begin{algorithm}
    \caption{Mapping Decision Algorithm}\label{algo:mappingdecision}
    \begin{algorithmic}[1]
    \Require $E^{t}$ and $V^{t}$
    \Ensure $P^{t}$
    \State $E_{inside}^{t} \leftarrow \emptyset$ \Comment Filtering out vehicles.
    \ForEach{$e \in E^{t}$}
        \If{$e.inside > Threshold_{inside}$}
            \State $E_{inside}^{t} \leftarrow E_{inside}^{t} \cup e$
        \EndIf
    \EndFor
    \ForEach{$e \in E_{inside}^{t}$} \Comment Score table.
        \ForEach{$v \in V^{t}$}
            \State $ST[e][v] \leftarrow Score(e,v)$
        \EndFor
    \EndFor
    \ForEach{$(e,v) \in ST$} \Comment Confidence table.
        \State $C[e][v] \leftarrow Confidence(e,v)$
    \EndFor
    \State $P^{t} \leftarrow \emptyset$ \Comment Mapping
    \While {True}
        \State $(max_e, max_v) \leftarrow FindMaximumPair(C)$
        \If{$Score(max_e, max_v) \neq 0$}
            \State $P^{t} \leftarrow P^{t} \cup (max_e, max_v)$
            \State Set all elements in the row $max_e$ to 0.
            \State Set all elements in the column $max_v$ to 0.
        \Else
            \State \textbf{\emph{break}}
        \EndIf
    \EndWhile
    \end{algorithmic}
\end{algorithm}

\begin{itemize}
    \item Filtering out vehicles: The goal of the 1-6 lines of the algorithm is to filter out the vehicles not in the image. In the BBX-MDFNN, each element $e$ in $E^{t}$ is a vector of length $5$, where the last element, $e.inside$, represents the probability of the vehicle being inside the image. Only vehicles with $e.inside$ greater than a threshold are considered for pairing, and $E_{inside}^t$ represents the set of vehicles considered for pairing.
    \item Creating the score table: To compare the similarity between $E_{inside}^t$ and $V^t$, we define a score function, where a higher score indicates greater similarity. In BBX-MDFNN, $e_j \in E^t$ represents the estimated bounding box with its first four elements, and $v_i \in V^t$ represents the bounding box itself. The score function is determined by calculating the Intersection over Union (IoU) and the center point distance, as expressed by the following formula:
    \begin{equation}\label{eq:bbx_score}
    \begin{aligned}
    Score_{bbx}(e_j,v_i) = (1-\omega)(\frac{Intersection(e_j,v_i)}{Union(e_j,v_i)}) +\\
    \omega(\frac{DiagonalLen-dist(e_j,v_i)}{DiagonalLen})\text{,}
    \end{aligned}
    \end{equation}
    where, $\omega$ is an adjustable weight, $Intersection(e_j,v_i)$ represents the intersection area of the bounding boxes $e_j$ and $v_i$, $Union(e_j, v_i)$ represents the union area of the bounding boxes $e_j$ and $v_i$, $DiagonalLen$ denotes the diagonal length of the image, and $dist(e_j, v_i)$ represents the center point distance between the bounding boxes $e_j$ and $v_i$.
    Based on the score function, we can establish a score table, as shown in Table~\ref{tab:scoretable}, represented by $ST$.
    \begin{table}
    \centering
    \caption{An example of the score table.}
    \begin{tabular}{| c | c | c | c |}
        \hline
        & $v_1$ & $v_2$ & $v_3$\\
        \hline
        $e_1$ & $0.3$ & $0.7$ & $0.1$\\
        \hline
        $e_2$ & $0.1$ & $0.83$ & $0.8$\\
        \hline
        $e_3$ & $0.62$ & $0.35$ & $0.4$\\
        \hline
    \end{tabular}
    \label{tab:scoretable}
    \end{table}
    \item Converting the score table to the confidence table: With the score table, we can now perform the mapping based on the scores. In the example from Table~\ref{tab:scoretable}, we would get $\{(e_2,v_2 ),(e_3,v_1 ),(e_1,v_3)\}$. However, this combination may not be the best, as the score for $(e_1,v_3)$ is only $0.1$. In fact, if $e_j$ assigns high scores to multiple bounding boxes simultaneously, it indicates that $e_j$ may not be confident about which bounding box is correct. Conversely, if $e_j$ assigns a high score to only one bounding box, it suggests that $e_j$ is confident about this choice. Therefore, $\{(e_1,v_2 ),(e_2,v_3 ),(e_3,v_1)\}$ could be a better combination. Consequently, we define a confidence function to establish a confidence table. The formula for the confidence function is as follows:
    \begin{equation}\label{eq:confidence}
    Confidence(e_j,v_i) = \frac{Score(e_j,v_i)}{\sum\limits_{k=1}^{|V^{t}|}Score(e_j,v_k)}\text{.}
    \end{equation}
    According to this confidence function, we can convert the score table in Table~\ref{tab:scoretable} into the confidence table as shown in Table~\ref{tab:confidencetable}.
    \begin{table}
    \centering
    \caption{An example of the confidence table.}
    \begin{tabular}{| c | c | c | c |}
        \hline
        & $v_1$ & $v_2$ & $v_3$\\
        \hline
        $e_1$ & $0.27$ & $0.64$ & $0.09$\\
        \hline
        $e_2$ & $0.06$ & $0.48$ & $0.46$\\
        \hline
        $e_3$ & $0.45$ & $0.26$ & $0.29$\\
        \hline
    \end{tabular}
    \label{tab:confidencetable}
    \end{table}
    \item Mapping: With the confidence table, we can now perform the mapping based on confidence values. Note that rows and columns that have been mapped will not participate in the subsequent mapping. For detailed specifics, please refer to lines 15-26 in Algorithm~\ref{algo:mappingdecision}.
\end{itemize}

Finally, it should be noted that for the paired vehicle $e_j$, the corresponding value of $v_i$ is input into the feedback layer of MDFNN at the next time step. Conversely, for vehicles that are not paired, the values are set to 0 and then input into the feedback layer of MDFNN.

\subsection{Federated Learning based MDFNN (FedMDFNN)}\label{subsec:fedmdfnn}

In our federated learning framework, each vehicle is treated as a federated client. After training its local MDFNN model, each vehicle uploads the MDFNN model parameters to the federated server. Upon receiving the model parameters from the participating vehicles, the federated server aggregates the model parameters and returns the aggregated parameters to the participating vehicles. This process enables the participating vehicles to update their local parameters. Subsequently, the participating vehicles continue training with the updated local parameters, and this process repeats. After a certain number of rounds, a satisfactory global model is generated. This trained global model is then returned to each participating vehicle and used for identifying the positions of vehicles in the images.

In the above process, participating vehicles do not need to transmit sensor values or image data to the server. They only need to send the parameters of their local MDFNN models. This not only ensures that sensitive data does not need to be uploaded, but also saves the network bandwidth required for uploading image data.

\section{Experiments} \label{sec:experiments}

\subsection{Simulation Environment and Dataset Generation}

We conducted experiments and validated our proposed method using the Carla simulator\cite{carla_Dosovitskiy17}. We utilized five maps provided by Carla for experimentation, namely Small Town, Middle Town, Highway, Square Town with Multiple Lanes, and Big City. Experiments were conducted on each map under 14 different weather conditions. A total of 100 vehicles were deployed on the roads, and these vehicles varied their speeds over time while adhering to traffic rules on the simulated maps (e.g., traffic lights). One of the vehicles, denoted as $x$, was equipped with front and rear cameras. All vehicles recorded their orientation, speed, and GPS coordinates. We assumed a sensor data acquisition rate of every $0.5$ seconds. In the end, we obtained a total of 3,500 front camera image data, 3,500 rear camera image data, and 350,000 sensor data entries. When vehicle $x$ received a message, it could calculate the distance to the sending vehicle, since the message contained the GPS information of the sending vehicle. Vehicle $x$ processed only messages from vehicles within a distance of 50 meters.

\begin{figure}
    \centering
    \begin{subfigure}[b]{0.38\textwidth}
        \centering
        \includegraphics[width=\textwidth]{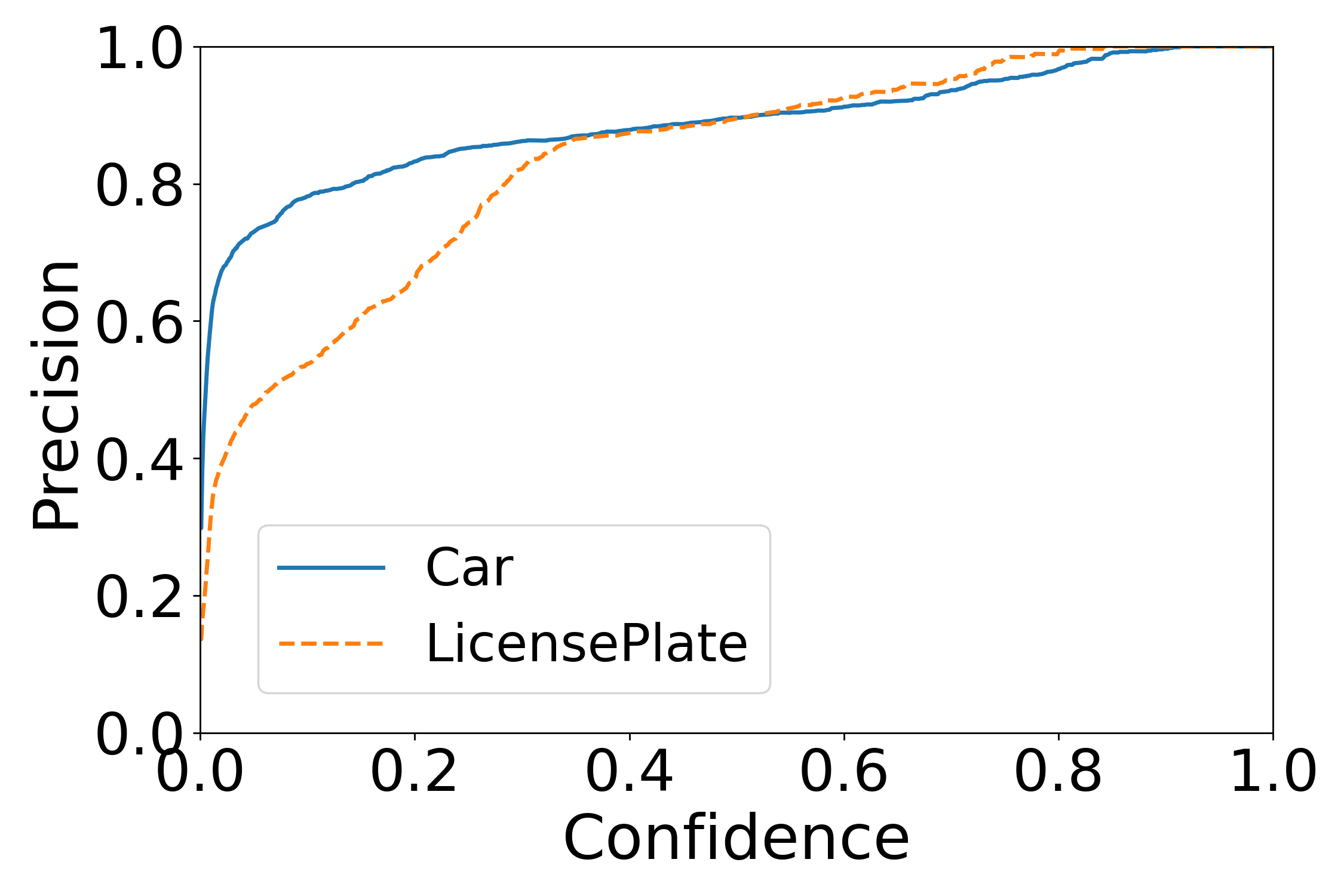}
        \caption{Precision}
    \end{subfigure}
    \begin{subfigure}[b]{0.38\textwidth}
        \centering
        \includegraphics[width=\textwidth]{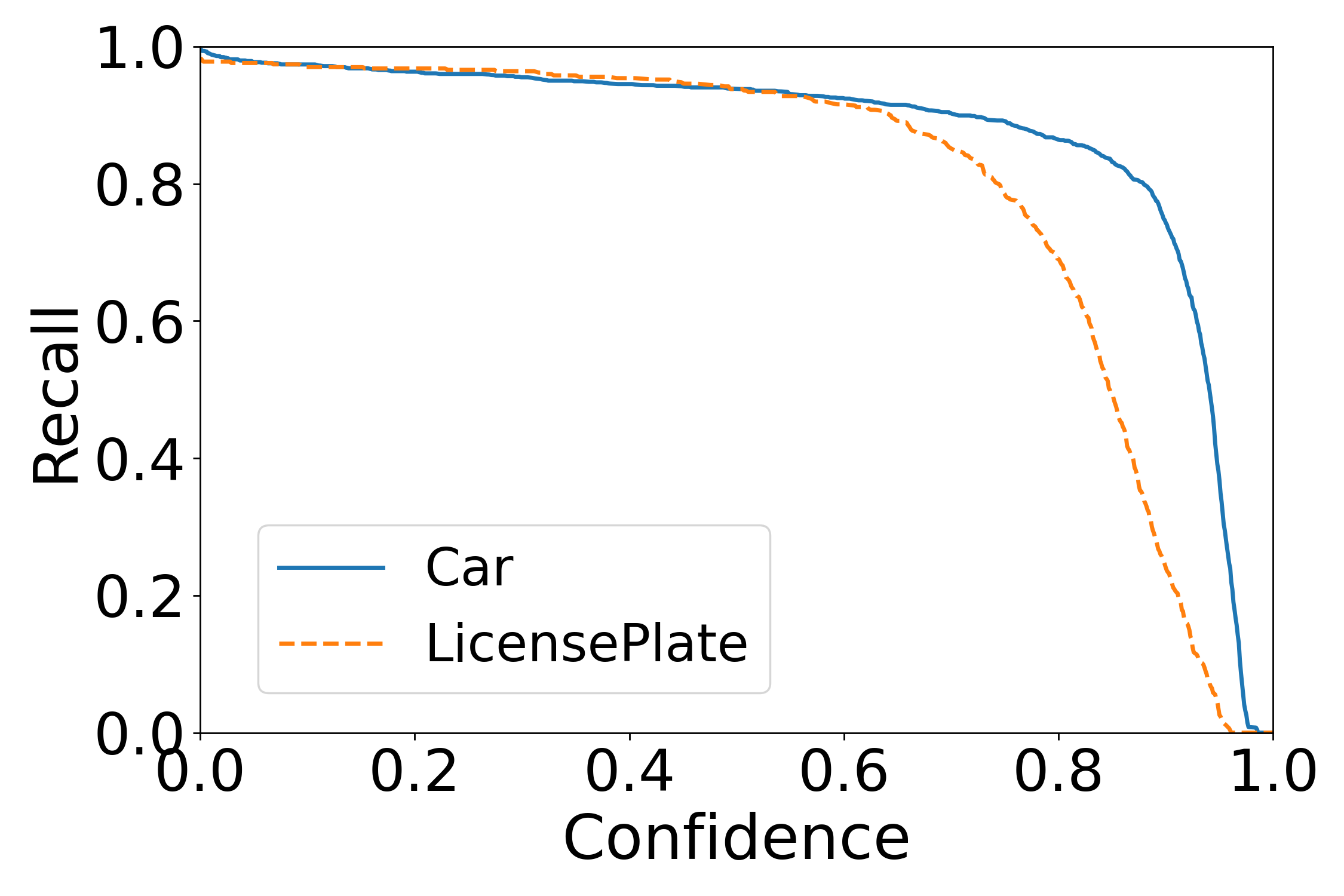}
        \caption{Recall}
    \end{subfigure}
    \caption{The performance of the retrained model.}
    \label{fig:yolo_modelperformance}
\end{figure}

During the experimentation, we utilized YOLOv5\cite{yolo_glenn_jocher_2020_4154370} to detect the bounding boxes of vehicles and license plates. The vehicle recognition model provided by YOLOv5 is trained for real-world vehicles. However, the vehicles in Carla differ slightly from their real-world counterparts. In addition, YOLOv5 does not offer a model for license plate recognition. Consequently, we retrained models for both vehicle and license plate recognition. The performance of our trained models is shown in Fig.~\ref{fig:yolo_modelperformance}. Fig.~\ref{fig:yolo_modelperformance}(a) illustrates the relationship between precision and confidence, while Fig.~\ref{fig:yolo_modelperformance}(b) shows the relationship between recall and confidence. We observed that as the confidence threshold increased, the recognition accuracy improved, but the number of successfully identified objects decreased. In other words, although precision could be enhanced, recall decreased significantly. Therefore, for subsequent experiments, we set the confidence threshold to 0.7 to strike a balance between precision and recall.

\subsection{Performance of the Automatic Labeling Submodule}\label{subsection:experiment_al}

Firstly, we are interested in the proportion of vehicles in a single image that can be labeled using the automatic labeling method proposed in this paper. If this proportion is too low, it implies that we may not be able to gather sufficient data for local training. Conversely, if this proportion is too high, it suggests that using the automatic labeling method alone is sufficient for pairing vehicles, and there might not be a need for FedMDFNN.

\begin{figure}
    \centering
    \includegraphics[width=8.8cm]{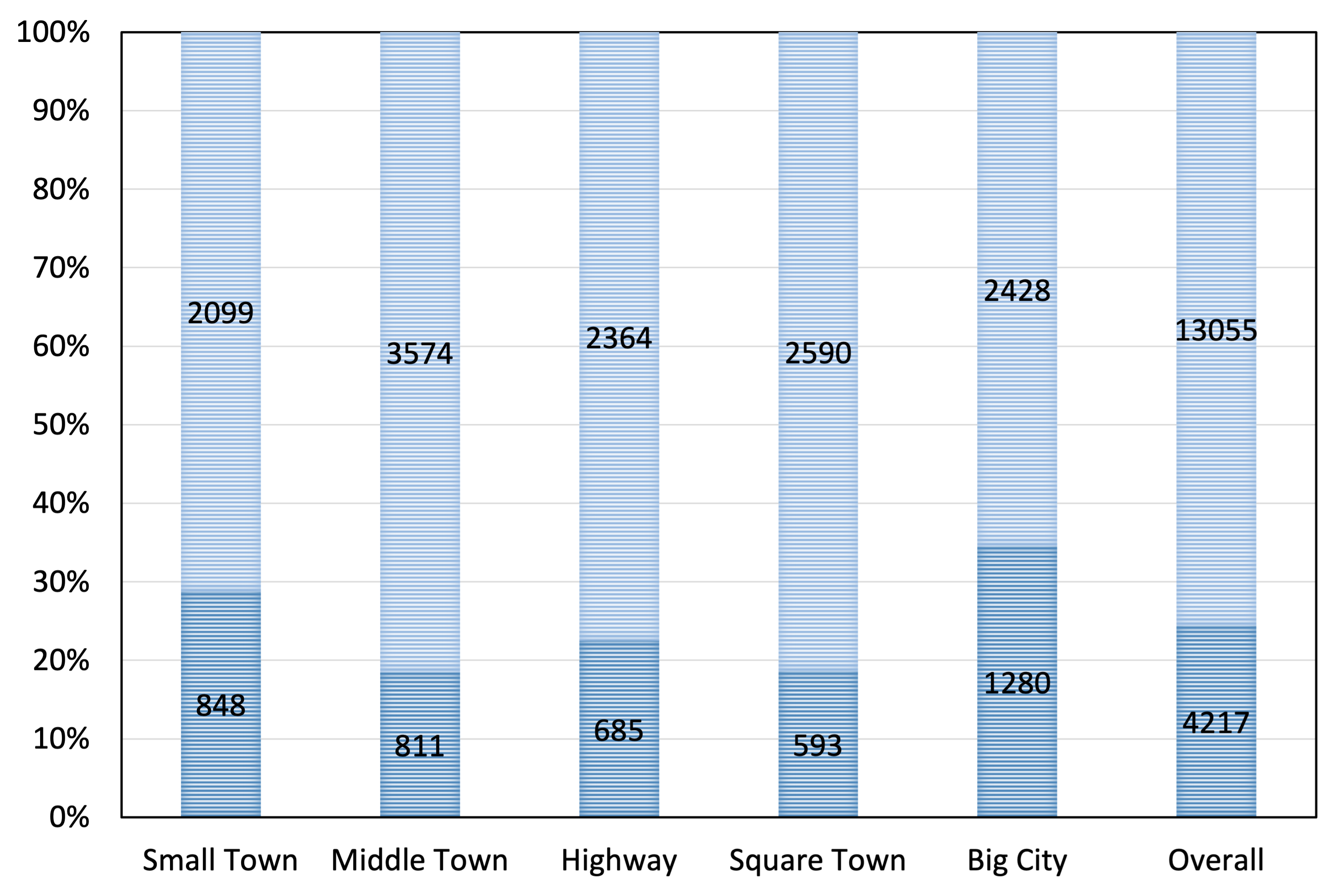}
    \caption{The performance of the automatic labeling submodel.}
    \label{fig:alsubmodule_1}
\end{figure}

Fig.~\ref{fig:alsubmodule_1} shows the results. In general, approximately 24\% vehicles can be paired using the automatic labeling method, while the remaining 76\% vehicles can be paired using FedMDFNN. Sect.~\ref{subsection:experiment_alda} shows that the data generated by the automatic labeling method are sufficient to train a good MDFNN model. Additionally, this experiment also shows that FedMDFNN is required to pair more vehicles.

We also conducted an experiment for the character conversion method proposed in Sect.~\ref{subsec:automaticlabeling}. Without character conversion, only about 16\% vehicles can be successfully paired. This indicates that the proposed character conversion indeed increases the proportion of paired vehicles.


The data of these successfully paired vehicles can serve as the training dataset required for training the MDFNN model. In the next section, these automatically labeled data are used to train MDFNN and validate the model performance.

\subsection{Performance of the Automatic Labeling and Data Augmentation Module}\label{subsection:experiment_alda}

In this experiment, three datasets are used separately to train the MDFNN model:
\begin{itemize}
    \item $\mathcal{D}_{AL}$: This dataset is generated using the automatic labeling with the front camera and data augmentation with the rear camera, meaning that it includes only the data from the front and rear cameras.
    \item $\mathcal{D}_{ALDA}$: This dataset is generated using the complete automatic labeling and data augmentation module.
    \item $\mathcal{D}_{Manual}$: The labeling data for this dataset are completed manually.
\end{itemize}

The three datasets only label data for Small Town, Middle Town, Highway, and Big City. $\mathcal{D}_{Manual}$ undoubtedly has the most labeled data, while the number of entries in the $\mathcal{D}_{AL}$ dataset is the lowest. The numbers of entries in the datasets are shown in Table~\ref{tab:datasets}. The models trained on these three datasets are validated using the Square Town map.

\begin{table}
    \centering
    \caption{Number of entries in dataset.}
    \begin{tabular}{| c | c | c | c |}
    \hline
    Dataset & Inside & Outside & Total\\
    \hline
    $\mathcal{D}_{AL}$ & 1734 & 1890 & 3624\\
    \hline
    $\mathcal{D}_{ALDA}$ & 1734 & 16171 & 17905\\
    \hline
    $\mathcal{D}_{Manual}$ & 7249 & 16956 & 24205\\
    \hline
    \end{tabular}
    \label{tab:datasets}
\end{table}

To compare the performance of these three datasets, we define $CR_{ic}$, $CR_{inside}$, $CR_{outside}$, and $CR_{total}$ as follows:
\begin{equation}\label{eq:cr_ic}
CR_{ic}=\frac{P_{correctly}}{N_{inside}}
\end{equation}
\begin{equation}\label{eq:cr_inside}
CR_{inside}=\frac{P_{inside}}{N_{inside}}
\end{equation}
\begin{equation}\label{eq:cr_outside}
CR_{outside}=\frac{P_{outside}}{N_{outside}}
\end{equation}
\begin{equation}\label{eq:cr_total}
CR_{total}=\frac{P_{correctly}+P_{outside}}{N_{inside}+N_{outside}}\text{,}
\end{equation}
where $P_{correctly}$ represents the number of vehicles correctly paired in the image by MDFNN; $N_{inside}$ represents the number of vehicles labeled as inside the image; $P_{inside}$ represents the number of vehicles that are actually inside the image, and identified as inside by MDFNN; $P_{outside}$ represents the number of vehicles actually outside the image, identified as outside the image by MDFNN; $N_{outside}$ represents the number of vehicles labeled as outside the image.

\begin{table}
    \centering
    \caption{The performance of Grid-MDFNN under different datasets.}
    \begin{tabular}{| c | c | c | c | c |}
    \hline
    Dataset & $CR_{ic}$ & $CR_{inside}$ & $CR_{outside}$ & $CR_{total}$\\
    \hline
    $\mathcal{D}_{AL}$ & 50.35\% & 65.23\% & 92.46\% & 76.76\%\\
    \hline
    $\mathcal{D}_{ALDA}$ & 53.42\% & 67.36\% & 94.14\% & 78.96\%\\
    \hline
    $\mathcal{D}_{Manual}$ & 55.84\% & 68.18\% & 94.21\% & 79.91\%\\
    \hline
    \end{tabular}
    \label{tab:gridmdfnn}
\end{table}

\begin{table}
    \centering
    \caption{The performance of BBX-MDFNN under different datasets.}
    \begin{tabular}{| c | c | c | c | c |}
    \hline
    Dataset & $CR_{ic}$ & $CR_{inside}$ & $CR_{outside}$ & $CR_{total}$\\
    \hline
    $\mathcal{D}_{AL}$ & 61.33\% & 77.74\% & 92.11\% & 80.63\%\\
    \hline
    $\mathcal{D}_{ALDA}$ & 65.29\% & 79.63\% & 94.42\% & 83.56\%\\
    \hline
    $\mathcal{D}_{Manual}$ & 72.08\% & 79.87\% & 93.68\% & 85.63\%\\
    \hline
    \end{tabular}
    \label{tab:bbxmdfnn}
\end{table}

The performance of Grid-MDFNN and BBX-MDFNN under different datasets is shown in Table~\ref{tab:bbxmdfnn} and Table~\ref{tab:gridmdfnn}. As expected, the dataset generated using manually labeled data performs the best, as it has the most data. However, models trained on datasets generated using automatic labeling with data augmentation still achieve good results. Compared to using manually labeled datasets, in BBX-MDFNN, $CR_{total}$ decreases from 85.63\% to 83.56\%, a slight decrease of 2.07\%.

We then analyze the reason for the lower $CR_{inside}$ compared to $CR_{outside}$. The vehicle recognition model that we used cannot identify all vehicles. According to Fig.~\ref{fig:yolo_modelperformance}(b), when we set the confidence value to 0.7, the recall value is 89.87\%, indicating that about 10\% of vehicles cannot be successfully identified by the vehicle recognition model (i.e., false negative cases). In cases where the number of bounding boxes is not sufficient, some vehicles cannot be paired, and thus, they are not counted in $P_{inside}$.

\begin{figure}
    \centering
    \begin{subfigure}[b]{0.23\textwidth}
        \centering
        \includegraphics[width=\textwidth]{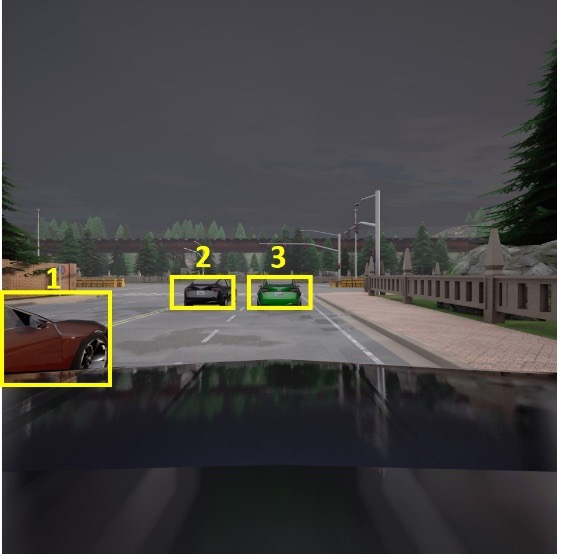}
        \caption{The case of easy to pair.}
    \end{subfigure}
    \begin{subfigure}[b]{0.23\textwidth}
        \centering
        \includegraphics[width=\textwidth]{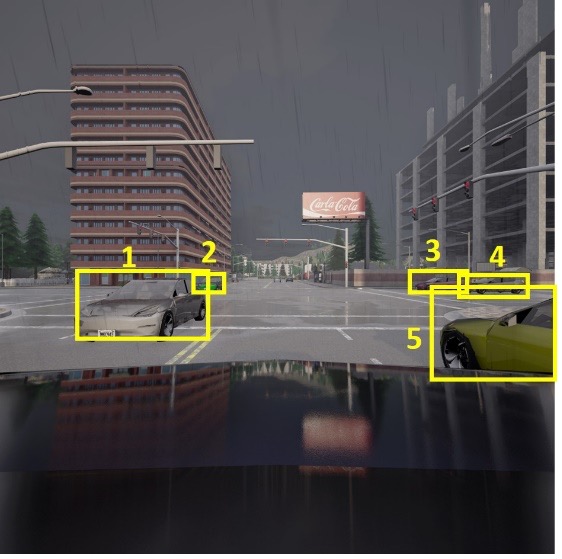}
        \caption{The case of hard to pair.}
    \end{subfigure}
    \caption{Analysis of pairing errors cases.}
    \label{fig:pairerroranalysis}
\end{figure}

We conducted further analysis on cases of pairing errors and found that weather is not the most crucial factor affecting pairing accuracy. The number of bounding boxes has a more significant impact. As the number of bounding boxes increases, it indicates that the bounding boxes are closer or even overlap, making pairing more challenging. Fig.\ref{fig:pairerroranalysis} illustrates an example. In Fig.\ref{fig:pairerroranalysis}(a), there are only three vehicles, and all three are correctly paired. However, in Fig.~\ref{fig:pairerroranalysis}(b), there are five vehicles. Due to the close proximity and overlapping bounding boxes of vehicles 1 and 2, the retrained vehicle recognition model only identifies the bounding box of vehicle 1. As a result, only vehicle 1 is correctly paired, while vehicles 3 and 4 are erroneously paired with each other's bounding boxes. Vehicle 5 is correctly paired.



\subsection{Performance Analysis of the FedMDFNN}\label{subsection:experiment_fedmdfnn}

In this section, we compare the performance of the federated FedMDNN with the centralized MDFNN. In the experiment, we used Flower\cite{flower2020} to implement the federated environment. We distributed the $\mathcal{D}_{ALDA}$ dataset to 2 client nodes. Each client locally trained its model and uploaded the model parameters to the federated server. The server then performed parameter aggregation. The final globally aggregated model was validated for performance using the Square Town map. Table~\ref{tab:fedmdfnn} presents the results.

\begin{table}
    \centering
    \caption{The performance of FedMDFNN}
    \begin{tabular}{| c | c | c | c | c |}
    \hline
    Model & $CR_{ic}$ & $CR_{inside}$ & $CR_{outside}$ & $CR_{total}$\\
    \hline
    Grid-FedMDFNN & 48.11\% & 64.46\% & 92.18\% & 75.75\%\\
    \hline
    Grid-MDFNN & 53.42\% & 67.36\% & 94.14\% & 78.96\%\\
    \hline
    BBX-FedMDFNN & 57.32\% & 75.27\% & 91.30\% & 78.63\%\\
    \hline
    BBX-MDFNN & 65.29\% & 79.63\% & 94.42\% & 83.56\%\\
    \hline
    \end{tabular}
    \label{tab:fedmdfnn}
\end{table}

From the experimental results, we observe that both BBX-FedMDFNN and Grid-FedMDFNN exhibit performance gaps compared to centralized BBX-MDFNN or Grid-MDFNN. In the case of Grid-FedMDFNN, $CR_{total}$ decreases from 78.96\% to 75.75\%, while in the case of BBX-FedMDFNN, $CR_{total}$ decreases from 83.56\% to 78.63\%, representing decreases of 3.21\% and 4.93\%, respectively. One reason for this is that in FedMDFNN, the amount of data used by each client to train the local model is only half of the data used to train centralized MDFNN. However, FedMDFNN ensures data privacy, coupled with the automatic labeling and data augmentation proposed in this paper (which is the reason for using the $\mathcal{D}_{ALDA}$ dataset in this experiment). Compared to centralized MDFNN, FedMDFNN demonstrates greater feasibility.

\section{Conclusions}\label{sec:conclusions}

This paper proposes FedMDFNN, which combines federated learning and automatic labeling to address the vehicle identification problem. Through the federated learning framework, drivers only need to upload local model parameters to aggregate and obtain the global model. This approach ensures user privacy as only parameters are uploaded, reducing network traffic consumption. Furthermore, we integrate automatic labeling and data augmentation to acquire the dataset needed for the FedMDFNN model, eliminating the need for manual labeling. This greatly enhances the overall feasibility of the approach, because typically drivers are reluctant to label data. In the future, we aim to explore the integration of object tracking technology to obtain more training data and enhance the model's performance. Additionally, as federated learning is extensively studied, we will also consider the common Non-IID issues in federated learning and their impact on the FedMDFNN.

\bibliographystyle{unsrtnat}
\bibliography{FedMDFNN}  






\end{document}